\documentclass{article}

\usepackage[preprint,nonatbib]{neurips_2021}

\usepackage[utf8]{inputenc} % allow utf-8 input
\usepackage[T1]{fontenc}    % use 8-bit T1 fonts
\usepackage{hyperref}       % hyperlinks
\usepackage{url}            % simple URL typesetting
\usepackage{booktabs}       % professional-quality tables
\usepackage{amsfonts}       % blackboard math symbols
\usepackage{nicefrac}       % compact symbols for 1/2, etc.
\usepackage{microtype}      % microtypography
\usepackage{xcolor}         % colors

\usepackage{amsmath,amsfonts,bm}

\def\eqref#1{equation~\ref{#1}}
\def\1{\bm{1}}

\def\vtheta{{\bm{\theta}}}

\DeclareMathAlphabet{\mathsfit}{\encodingdefault}{\sfdefault}{m}{sl}
\SetMathAlphabet{\mathsfit}{bold}{\encodingdefault}{\sfdefault}{bx}{n}

\usepackage{graphicx}
\usepackage{comment}
\usepackage{amsmath,amssymb} % define this before the line numbering.
\usepackage{color}
\usepackage{makecell}
\usepackage{array}
\newcolumntype{C}[1]{>{\centering\arraybackslash}m{#1}}
\newcolumntype{R}[1]{>{\raggedleft\arraybackslash}m{#1}}
\newcolumntype{P}[1]{>{\raggedright\arraybackslash}p{#1}}
\newcolumntype{M}[1]{>{\centering\arraybackslash}m{#1}}
\usepackage{multirow}
\usepackage{algorithm}
\usepackage{algorithmic}
\floatstyle{ruled}
\restylefloat{algorithm}
\usepackage[font=small,labelfont=bf,labelsep=period]{caption}
\usepackage{wrapfig}

\usepackage{xcolor}

\newcommand{\etal}{\textit{et al}.}
\newcommand{\ie}{\textit{i}.\textit{e}., }
\newcommand{\eg}{\textit{e}.\textit{g}., }

\title{Uncertainty-aware Clustering for Unsupervised Domain Adaptive Object Re-identification}

\author{{Pengfei Wang{$^{1}$}}\quad Changxing Ding{$^{1} $}\thanks{Corresponding Author.} \quad Wentao Tan{$^{1}$} \quad Mingming Gong{$^{2}$} \quad Kui Jia{$^{1}$} \quad  Dacheng Tao{$^{3}$}\\
	$^{1}$\	South China University of Technology \\ ~~$^{2}$\, University of Melbourne\\
	$^{3}$\ JD Explore Academy, China \\
	{\tt\small \{eepengfei.wang, eewentaotan\}@mail.scut.edu.cn, \{chxding, kuijia\}@scut.edu.cn}, \\\tt\small {mingming.gong@unimelb.edu.au, taodacheng@jd.com}
	}

\begin{document}

\maketitle

\begin{abstract}
Unsupervised Domain Adaptive (UDA) object re-identification (Re-ID) aims at adapting a model trained on a labeled source domain to an unlabeled target domain. State-of-the-art object Re-ID approaches adopt clustering algorithms to generate pseudo-labels for the unlabeled target domain. However, the inevitable label noise caused by the clustering procedure significantly degrades the discriminative power of Re-ID model. To address this problem, we propose an uncertainty-aware clustering framework (UCF) for  UDA tasks. First, a novel hierarchical clustering scheme is proposed to promote clustering quality. Second, an uncertainty-aware collaborative instance selection method is introduced to select images with reliable labels for model training. Combining both techniques effectively reduces the impact of noisy labels. In addition, we introduce a strong baseline that features a compact contrastive loss. Our UCF method consistently achieves state-of-the-art performance in multiple UDA tasks for object Re-ID, and significantly reduces the gap between unsupervised and supervised Re-ID performance.
%  In particular, our UCF method achieves better performance than the fully supervised setting in the MSMT17$\to$Market1501 task.
 In particular, the performance of our unsupervised UCF method in the MSMT17$\to$Market1501 task is better than that of the fully supervised setting on Market1501. The code of UCF is available at \url{https://github.com/Wang-pengfei/UCF}.

\end{abstract}

\section{Introduction}
\label{introduction}

The goal of object re-identification (Re-ID) is to retrieve object images belonging to the same identity across different camera views. Due to its broad range of potential applications, (\eg smart retail), Re-ID research has experienced explosive growth in recent years \cite{11_zhang2019densely, 06_li2018harmonious, sun2018beyond, 18_xu2018attention,luo2019strong,wei2018glad,ding2020multi,yan2021beyond,jiang2021ph,wang2021batch,gong2021lag,wan2019concentrated,zhao2020deep,zhang2020learning}. Most existing approaches achieve remarkable performance when the training and testing data are drawn from the same domain. However, due to the presence of significant domain gaps, Re-ID models trained on source datasets typically exhibit clear performance drops when directly applied to the target datasets. Unsupervised Domain Adaptive (UDA) object Re-ID is therefore proposed to adapt the model trained on the source image domain with identity labels to the target image domain without the need for identity annotations. Unlike the traditional UDA setting, which assumes that both domains share the same classes, UDA in object Re-ID is a more challenging open-set problem, in that the two domains have totally different identities (classes).

State-of-the-art methods \cite{song2018unsupervised,ge2020mutual,zhai2020ad,yang2019selfsimilarity,zhang2019self,wang2020unsupervised,zhao2020unsupervised,li2020joint,ge2020selfpaced} adopt clustering algorithms to generate pseudo-labels for the target domain. At the beginning of each epoch, a clustering algorithm is applied on the features extracted from the current model to generate pseudo-labels for each image. The current model is then updated via retraining with the pseudo-labels. These two steps alternate so that the model gradually adapts to the target data. While pseudo-label approaches have achieved promising results, there are still two major challenges to deal with. First, due to the domain gap, the current model is not an optimal feature extractor for the target domain; second, the unsupervised nature of the clustering makes it difficult to obtain the real identity labels, even given the optimal feature extractor. The obtained pseudo-labels therefore usually contain a certain level of noise, which undermines the final Re-ID performance.

In this paper, we propose an uncertainty-aware clustering framework (UCF) to handle the above problem from two perspectives. First, we identify and decompose unreliable clusters using a novel hierarchical clustering algorithm. Due to the domain shift, the Re-ID model has limited discriminative power in the target domain; as a result, inter-class distances may vary dramatically. This means that images of visually similar identities may be grouped into the same cluster, the size of which tends to be large. To handle this problem, we first adopt a clustering algorithm, such as DBSCAN~\cite{ester1996density}, to perform coarse clustering. We then calculate the silhouette coefficiency~\cite{Rousseeuw}, which measures both the tightness and separation of each cluster. For clusters with small silhouette coefficiency, we further perform fine-grained clustering within the cluster. In this way, unreliable clusters can be decomposed into several smaller ones.

Second, we identify images with unreliable pseudo-labels using a novel uncertainty-aware collaborative instance selection method. Specifically, we adopt a deep network and its temporally averaged model, \ie the mean-Net~\cite{tarvainen2017mean}, to cluster images in the target domain, respectively. Since these two models have different learning capabilities, their clustering results will be different. We then evaluate whether each instance is located in similar clusters across the two networks. If a large number of overlapping samples exist in the two clusters, the clustering result of this instance is considered to be reliable. Finally, we only adopt instances with reliable labels for model training, which reduces the impact of noise in the pseudo-labels.

Through joint hierarchical clustering and reliable sample selection, our UCF framework can effectively reduce the adverse effects of noisy pseudo-labels. We further propose a compact contrastive loss for UDA Re-ID. Recent approaches~\cite{ge2020selfpaced,zhong2019invariance,li2020joint} typically adopt contrastive loss for model training. However, these losses require all image features in the target domain to be stored in the memory bank. This may result in two problems: first, this strategy consumes a lot of memory; second, only the features of a small number of images are updated in each iteration. These problems become especially serious for large-scale Re-ID datasets, such as MSMT17~\cite{wei2018person}. To solve the above mentioned problems, we propose an improved contrastive loss using a class-level memory bank, which stores one single feature vector for each class rather than the features of all images.

Our main contributions can be summarized as follows: 1) We propose a strong baseline that adopts an improved contrastive loss using compact class-level memory banks; 2) We design a hierarchical clustering scheme to improve the quality of clustering, which decomposes unreliable clusters from coarse to fine; 3) We introduce a novel collaborative clustering method to identify images with unreliable pseudo-labels, which significantly relieves the impact of noise in pseudo-labels; 4) Our approach outperforms state-of-the-art methods by large margins on many UDA tasks for Re-ID.

\section{Related Works}
We review the literature in three parts: 1) unsupervised domain adaptive (UDA) object Re-ID, 2) contrastive learning, and 3) deep learning with noisy labels.

\subsection{UDA Object Re-ID}
% \paragraph{Unsupervised domain adaptive (UDA) object Re-ID}
Existing UDA approaches for object Re-ID can be roughly divided into two categories: pseudo-label-based methods~\cite{song2018unsupervised,yang2019selfsimilarity,zhang2019self,ge2020mutual,zhai2020ad,zhong2019invariance,yu2019unsupervised,wang2020unsupervised} and domain translation-based methods~\cite{deng2018image,wei2018person,chen2019instance,ge2020structured}. Domain translation-based methods transfer labeled images in the source domain to the style of the target domain images, then use these transferred images and the inherited ground-truth labels for model training. However, a gap inevitably arises between the translated image and the real target domain image, which affects the performance of these approaches. Pseudo-label-based methods group unannotated images using clustering algorithms and then train the network with pseudo-labels generated by clustering. For example, Li \etal \cite{li2020joint} employed both visual and temporal similarity cues to promote the quality of pseudo-labels. However, existing approaches typically ignore the noise remaining in pseudo-labels.

\begin{figure}[t]
\centering
\includegraphics[width=0.8\linewidth]{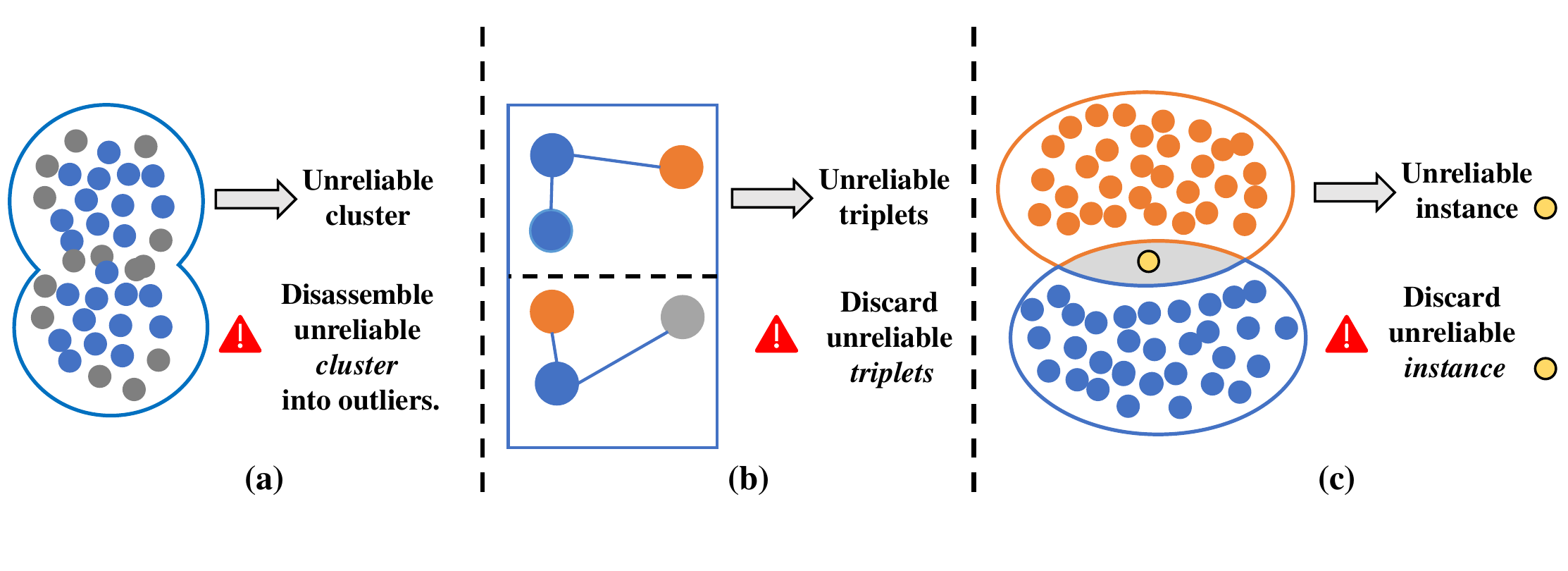}
\caption{(a) SpCL~\cite{ge2020mutual} regards all instances in one unreliable cluster as outliers.
(b) NRMT~\cite{zhao2020unsupervised} removes unreliable triplets. (c) Our UCF measures the uncertainty of each instance, which is more fine-grained. (Best viewed in color.)}
\label{fig:difference}
\end{figure}

Recently, some methods have been proposed that attempt to solve the label noise problem. Ge \etal~\cite{ge2020mutual} proposed generating more robust soft labels via mutual mean-teaching. However, the classifier trained with noisy labels forms the foundation for soft label generation, which hinders the improvement of Re-ID performance. Ge \etal~\cite{ge2020selfpaced} further proposed the SpCL approach. As shown in Fig. \ref{fig:difference}(a), it identifies and regards all instances in one unreliable cluster as outliers with reference to their proposed reliability criterion. However, removing all images in a cluster may waste samples with reliable pseudo-labels. Similarly, Zhao \etal~\cite{zhao2020unsupervised} introduced the Noise Resistible Mutual-Training (NRMT) approach, as shown in Fig. \ref{fig:difference}(b), which removes triplets that are considered to be unreliable. The reliability of a triplet is measured with reference to the distance between the features of the triplet samples extracted by two networks. Unlike the above works, our approach is more fine-grained, as it first improves the clustering quality and then removes unreliable instances rather than complete clusters or triplets.

% \begin{wrapfigure}{r}{0.5\linewidth}
% \centering
% \includegraphics[width=1.0\linewidth]{difference.pdf}
% \caption{(a) SpCL~\cite{ge2020mutual} regards all instances in one unreliable cluster as outliers.
% (b) NRMT~\cite{zhao2020unsupervised} removes unreliable triplets. (c) Our UCF measures the uncertainty of each instance, which is more fine-grained. (Best viewed in color.)}
% \label{fig:difference}
% \end{wrapfigure}

% \paragraph{Contrastive learning}
\subsection{Contrastive Learning}
As a promising paradigm of unsupervised learning, contrastive learning has lately achieved state-of-the-art performance in unsupervised  visual  representation  learning.
 Recently, contrastive learning methods combined with data augmentation strategies achieved great successes, such as SimCLR~\cite{chen2020simple}, MoCo~\cite{he2019momentum}, and BYOL~\cite{grill2020bootstrap}.  These methods treat each instance as a class represented by a feature vector and data pairs are constructed through data augmentations.
These methods treat each instance as a class, which yields poor results for the domain adaptive object Re-ID task, because the intra- and inter-class similarity on the unlabeled target domain cannot be measured accurately.
Some recent works~\cite{wang2020unsupervised,ge2020selfpaced,li2020joint,zhong2019invariance} have introduced improved contrastive loss to domain adaptation. For example, the SpCL approach~\cite{ge2020selfpaced} includes a unified contrastive loss, which jointly distinguishes source-domain classes, target-domain clusters, and un-clustered instances.
One common drawback of these methods is the need to store all instance features, which requires a large amount of memory.
 To solve this problem, we propose a new contrastive loss with a compact class-level memory bank, which resolves these issues by storing a single feature vector for each cluster rather than all instance features.

%\paragraph{Deep learning with noisy labels}
\subsection{Deep Learning with Noisy Labels}

Many studies have attempted to effectively train deep neural networks in the presence of noisy labels for close-set classification problems.
Some recent works have introduced a sample selection approach that selects data with reliable labels for training~\cite{lyu2019curriculum,malach2017decoupling}.
Notably, the small loss trick, which regards samples with small training loss as clean, has demonstrated powerful ability.
However, the small loss trick is not suitable to select clean samples in UDA object Re-ID task.
This is because the number of target domain clusters (classes) changes through re-clustering during the training process.
Moreover, recent studies suggest various ways in which additional performance gain can be achieved by maintaining two networks to avoid accumulating sampling bias~\cite{han2018co,yu2019does}.
For example, Co-teaching~\cite{han2018co} works by training two deep models simultaneously, where each network selects the small-loss instances as reliable samples for the other one.
These methods focus primarily on the close-set problems with pre-defined classes, which cannot be generalized to our open-set object Re-ID task with completely unknown classes on the target domain.

\section{Methodology}
\label{Methodology}
In this section, we present the details of our uncertainty-aware clustering framework (UCF), which reduces the effects of the noisy pseudo-labels in clustering-based Unsupervised Domain Adaptation (UDA). Our key idea is to select samples with reliable pseudo-labels in the target domain for model training purposes. To this end, we propose hierarchical clustering and uncertainty-aware collaborative instance selection methods to reduce the adverse effects of noisy pseudo-labels, and therefore improves the ability of model to learn cross-domain discriminative representations. In addition, we propose a strong baseline with a new contrastive loss using compact class-level memory banks.

Formally, we denote the source domain data as $\mathbb{D}_{s}=\{\left.\left(\boldsymbol{x}_{i}^{s}, {y}_{i}^{s}\right)\right|_{i=1} ^{N_{s}}\}$, where $\boldsymbol{x}_{i}^{s}$ and ${y}_{i}^{s}$ denote the $i$-th training instance and its annotation, respectively. The target-domain data without ground-truth labels are denoted as $\mathbb{D}_{t}=\{\left.\boldsymbol{x}_{i}\right|_{i=1} ^{N_{t}}\}$.  $N_{s}$ and $N_{t}$ denote the sample size in the source and target domains, respectively.

\begin{figure*}[t]
\centering
\includegraphics[width=1.0\linewidth]{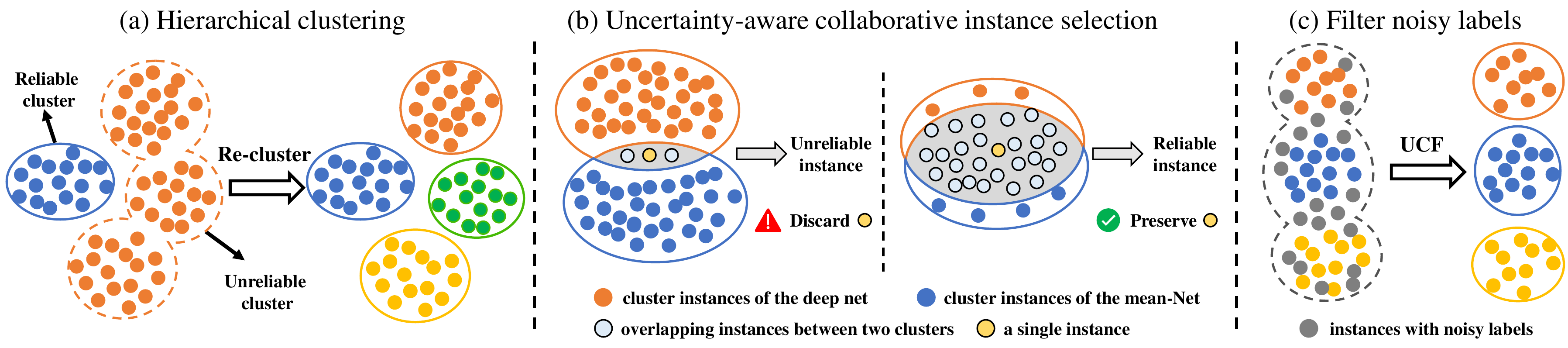}
\caption{(a) Illustration of the proposed hierarchical clustering (HC) method.
(b) Illustration of the proposed uncertainty-aware collaborative instance selection (UCIS) method. (c) Illustration of the overall clustering results by UCF. (Best viewed in color.)}
\label{fig:UCF}
\end{figure*}

\subsection{Supervised Pre-training for Source Domain}
\label{pretrain}
% \vspace{-5pt}
In the first stage of UCF, we train the Re-ID model $F(\cdot | \boldsymbol{\theta})$  with the labeled
source dataset $\mathbb{D}_{s}$ using the cross-entropy loss and the triplet loss
\cite{39hermans2017defense};
here, $\vtheta$ denotes parameters of the deep network.
The pre-trained Re-ID model has the basic
discriminability for domain adaptation.
We then adopt this pre-trained network $F(\cdot | \boldsymbol{\theta})$  to extract the features of the target domain images.
Following the existing clustering-based UDA methods \cite{zhang2019self,ge2020selfpaced,li2020joint},
% we use DBSCAN~\cite{ester1996density} to cluster the extracted features before each epoch.
% We use Jaccard distance to perform DBSCAN clustering on the target domain and obtain $K$ clusters.
we  use  DBSCAN~\cite{ester1996density} and  Jaccard distance to cluster the extracted features into $K$ clusters before each epoch.
We consider each cluster as a class and assign the same pseudo label for the instances belonging to the same cluster.
	
% \subsection{Reliable Sample Selection by Collaborative Clustering}
\subsection{Uncertainty-aware Clustering Framework}
\label{sec:general-cluster}

\begin{figure*}[t]
\centering
\includegraphics[width=1.0\linewidth]{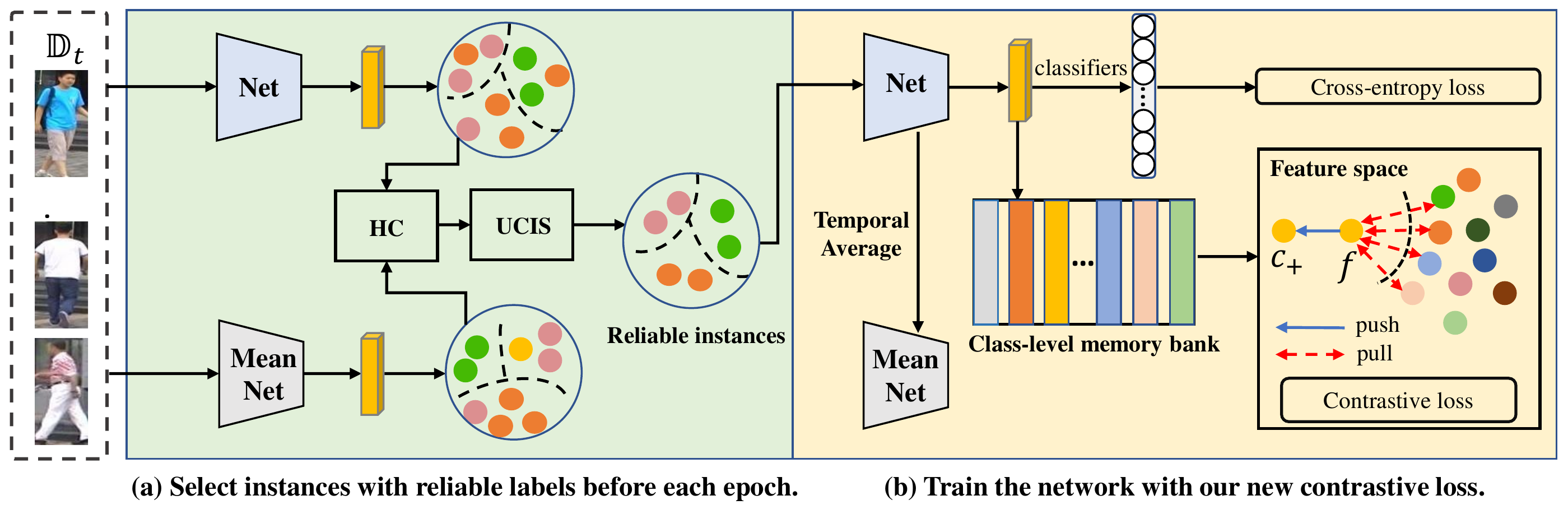}
\caption{Model architecture of UCF in the training stage.
UCF adopts a deep network and its temporally averaged model, \ie the mean-Net, to cluster images in the target domain.
 After that, the proposed novel hierarchical clustering scheme and the uncertainty-aware collaborative instance selection method are used to select images with reliable labels for model training.
This effectively reduces the impact of noisy labels. Step (a) and step (b) are performed alternately.
Note that the parameters of mean-Net model are not updated during back propagation.
In the testing stage, the mean-Net model is adopted for inference. (Best viewed in color.)}
%\vspace{-10pt}
\label{fig:net}
\end{figure*}

% \vspace{-5pt}
\paragraph{Hierarchical clustering}

As explained in Section \ref{introduction}, images of visually similar identities may be grouped into the same cluster, which introduces significant noise to the pseudo-labels. Recent method~\cite{ge2020selfpaced} simply regards all instances in the unreliable clusters as outliers. However, this strategy may result in a large number of informative instances being lost. In the following, we handle this problem using a hierarchical clustering (HC) method that conducts fine-grained clustering in these clusters.

Intuitively, a reliable cluster should be compact and independent from other clusters.
This means that the distances between instances in the same cluster should be small, and the distance between different clusters should be large.
To measure the reliability of one cluster, we first calculate the silhouette coefficiency~\cite{Rousseeuw} for each of its instances. Specifically, the silhouette coefficiency for the $i$-th instance in the $k$-th cluster is formulated as follows:

\begin{equation}
    \mathcal{S}(\boldsymbol{f}_k^i)=\frac{{b}(\boldsymbol{f}_k^i) - {a}(\boldsymbol{f}_k^i)}{\max({a}(\boldsymbol{f}_k^i), {b}(\boldsymbol{f}_k^i))} \in [-1,1],
\end{equation}
where $\boldsymbol{f}_k^i$ denotes the feature of the instance. ${a}(\boldsymbol{f}_k^i)$ represents the average distance between the $i$-th instance and all the other instances in the $k$-th cluster. Moreover, ${b}(\boldsymbol{f}_k^i)$ represents the average distance between the instance and all instances in the nearest cluster, which can be calculated as follows:
	\begin{equation}
	{a}(\boldsymbol{f}_k^i) = \frac{1}{|\mathcal{I}_k|-1}\sum_{}d_{J}(\boldsymbol{f}_k^i,\boldsymbol{f}_k^j), \boldsymbol{f}_k^j \in \mathcal{I}_k,i\neq j,
\end{equation}

	\begin{equation}
    	{b}(\boldsymbol{f}_k^i) =\min_{k\neq l} \{\frac{1}{|\mathcal{I}_l|}\sum_{}d_{J}(\boldsymbol{f}_k^i,\boldsymbol{f}_l^j)\}, \boldsymbol{f}_l^j \in \mathcal{I}_l,
\end{equation}
% After obtaining the silhouette coefficiency of each sample, we calculate the average silhouette coefficiency of each cluster:
where $d_{J}(\cdot,\cdot)$ represents the Jaccard distance, ${\cal I}_{k}$ (${\cal I}_{l}$) denotes the set of samples belonging to the $k$($l$)-th cluster, and $|\cdot|$ denotes the number of features in a cluster. Since the Jaccard distance between each pair of samples has been calculated during DBSCAN clustering, this step hardly increases time consumption.
Finally, we calculate the average silhouette coefficiency for the $k$-th cluster:
	\begin{equation}
	\mathcal{S}(\mathcal{I}_k) =\frac{1}{|\mathcal{I}_k|}\sum_{} \mathcal{S}(\boldsymbol{f}_k^i), \boldsymbol{f}_k^i \in \mathcal{I}_k. \label{eq:silhouette}
\end{equation}
When $\mathcal{S}(\mathcal{I}_k)<0$, the intra-class distance surpass the inter-class distance.
This usually indicates unreliable clustering from an object Re-ID perspective.
We adopt a threshold of $\alpha$ to select these unreliable clusters. As shown in Fig. \ref{fig:UCF}(a), we do not change the reliable ($\mathcal{S}(\mathcal{I}_k)>\alpha$) clusters, but we decompose an unreliable cluster into several smaller ones. In more detail, we use DBSCAN with the maximum neighbor distance $d$ for coarse clustering and then measure the reliability of each cluster. Within each unreliable cluster, we use DBSCAN with the maximum neighbor distance of $2/3d$ for fine-grained clustering. Since the number of samples in each unreliable cluster is limited, this step only adds a small amount of time consumption.

\paragraph{Uncertainty-aware collaborative instance selection}
\label{sec:Uncertainty}
Although hierarchical clustering improves the quality of clustering, there are still inevitably noisy pseudo-labels in many clusters.
In order to identify individual instances with noisy pseudo-labels, we propose an uncertainty-aware collaborative instance selection (UCIS) method, which adopts a deep network and its temporally averaged (mean-Net)~\cite{tarvainen2017mean} model to cluster the samples in the target domain separately. The parameters of the two models at iteration $T$ are denoted as $\boldsymbol{\theta}$ and $E^{(T)}[\boldsymbol{\theta}]$, respectively. $E^{(T)}[\boldsymbol{\theta}]$ is obtained as follows:
%\vspace{-5pt}
\begin{equation}
\label{eq:ema}
E^{(T)}[\boldsymbol{\theta}]  = \sigma E^{(T-1)}[\boldsymbol{\theta}] +(1-\sigma) \boldsymbol{\theta},
\end{equation}
where $\sigma$ is a temporal ensemble momentum coefficient whose value is within the range $[0,1)$.

After hierarchical clustering, we obtain the fine-grained clustering results of the two models.
We regard the clustering result of one instance as reliable if it is located in two similar clusters across the two models. The similarity of the two clusters is evaluated according to their overlap. More specifically, we propose the following metric to measure the clustering uncertainty of one instance $\boldsymbol{x}_i$:
\begin{equation}
    \mathcal{U}(\boldsymbol{x}_i)=\frac{|\mathcal{I}_k \cap {\mathcal{I}_{\text{mean}}}_l|}{|\mathcal{I}_k|} \in [0,1],
\end{equation}
where  $\mathcal{I}_k$ and ${\mathcal{I}_{\text{mean}}}_l$ denote the clusters containing $\boldsymbol{x}_i$ by the two models, respectively.
A larger $\mathcal{U}(\boldsymbol{x}_i)$ indicates larger overlap between $\mathcal{I}_k$ and ${\mathcal{I}_{\text{mean}}}_l$.

The value of $\mathcal{U}(\boldsymbol{x}_i)$ is therefore able to reflect the reliability of the pseudo-label for $\boldsymbol{x}_i$.
We set $\beta \in [0,1]$ as a threshold to select instances with reliable pseudo labels. As shown in Fig. \ref{fig:UCF}(b), in each epoch, we only preserve instances for model training where $\mathcal{U}(\boldsymbol{x}_i)$ is larger than $\beta$.

Here we adopt mean-Net to select instances with reliable pseudo labels in offline clustering.
Some methods train two networks together for close-set UDA problems~\cite{han2018co}, where the two networks select reliable samples for each other.
This strategy may not work well in our framework. This is because UCF selects samples deemed reliable by both networks based on uncertainty, which requires the two networks to have different discriminative power. However, as empirically proved in~\cite{ge2020mutual}, the two networks will obtain similar discriminative power if they are trained with exactly the same supervision signals. Therefore, mean-Net is a better choice in our framework.

% \vspace{-5pt}
\subsection{A Strong Baseline for Clustering-based UDA}
\label{sec:baseline}

Fig. \ref{fig:net} illustrates the structure of our method. Aside from the commonly used cross-entropy loss, we propose to use the following contrastive loss.
Given the feature $\boldsymbol{f}$ of one target domain instance, our proposed contrastive loss is formulated as follows:
\begin{equation}\label{eq:loss_contrastive}
    \mathcal{L}_c^t(\boldsymbol{\theta}) = -\log\frac{exp(\langle \boldsymbol{f}, \boldsymbol{c^{+}} \rangle /\tau)}{\sum_{k=1}^{K}{exp(\langle \boldsymbol{f}, \boldsymbol{c}_k \rangle /\tau)}},
\end{equation}
where $\boldsymbol{c^{+}}$ stands for the positive class prototype corresponding to $\boldsymbol{f}$, $\tau$ is  a temperature factor, and $\langle\cdot,\cdot\rangle$  denotes the inner product between two feature vectors. The loss value is low when $\boldsymbol{f}$ is similar to $\boldsymbol{c^{+}}$ and dissimilar to all the other cluster prototypes.

\paragraph{Memory initialization}
Each cluster is regarded as one class. The class-level memory bank $\{\boldsymbol{c_1}, \cdots, \boldsymbol{c}_{K}\}$ is initialized with the mean feature of each cluster. Formally,
\begin{equation}
	\boldsymbol{c}_{k} = \frac{1}{|{\cal I}_k|}\sum_{\boldsymbol{f}_k^i \in {\cal I}_{k}} \boldsymbol{f}_k^i. \label{eq:centroid}
\end{equation}

% \vspace{-5pt}
\paragraph{Memory update}

During training, $\boldsymbol{c}_k$ is continuously updated within each epoch, according to all instances in the $k$-th cluster:
	\begin{equation}
\label{eq:update}
    \boldsymbol{c}_{k} \gets m^t \boldsymbol{c}_{k} + (1-m^t)\boldsymbol{f}_k^i,  \boldsymbol{f}_k^i \in {\cal I}_{k},
\end{equation}
where $m^t \in [0,1]$ is the momentum coefficient for updating the target-domain class prototypes.

% \vspace{-5pt}
\paragraph{Discussion}

Compared with existing methods \cite{wu2018unsupervised,he2019momentum,chen2020simple,oord2018representation}, our proposed contrastive loss has two advantages. First, we only need to store class prototypes in the memory rather than the features of all samples, meaning that our approach has less memory cost. Second, each feature in the memory bank can be updated frequently within one epoch, which enables accurate loss computation in Eq. \ref{eq:loss_contrastive}.
In comparison, each feature in an instance-level memory bank can be updated only once per epoch, which may bring error in contrastive loss computation.

\section{Experiments}
% \vspace{-5pt}

\subsection{Datasets and Evaluation Protocol}
% \vspace{-5pt}

\begin{table*}[ht]
	\scriptsize
    % \footnotesize
    % \small
    % \tiny
    \caption{Statistics of the datasets used for training and evaluation}
	\label{tab:dataset}
	\centering
	\begin{tabular}{P{2.2cm}|C{0.9cm}C{0.9cm}C{0.9cm}C{0.9cm}C{0.9cm}C{0.9cm}C{0.9cm}}
	    \hline
    Dataset & \# type & \# train IDs & \# train images & \# test IDs & \# query images & \# cameras & \# total images \\
    \hline
    Market-1501~\cite{market} & real & 751 & 12,936 & 750 & 3,368 & 6 & 32,217 \\
    DukeMTMC-ReID~\cite{dukemtmc} & real & 702 & 16,522 & 702 & 2,228 & 8 & 36,411 \\
    MSMT17~\cite{wei2018person} & real & 1,041 & 32,621 & 3,060 & 11,659 & 15 & 126,441 \\
    PersonX~\cite{sun2019dissecting}  & synthetic & 410 & 9,840 & 856 & 5,136 & 6 & 45,792 \\
    % PersonX*~\cite{sun2019dissecting} & 410 & 9,840 & - & - & 6 & 9,840 \\
    \hline
    VeRi-776~\cite{liu2016deep} & real & 575 & 37,746 & 200 & 1,678 & 20 & 51,003 \\
    VehicleID~\cite{liuhy2016deep} & real & 13,164 & 113,346 & 800 & 5,693 & - & 221,763  \\
    % VehicleID~\cite{liuhy2016deep} & 13,164 & 113,346 & - & - & - & 113,346  \\
    VehicleX~\cite{naphade20204th}  & synthetic & 1,362 & 192,150 & - & - & 11 & 192,150 \\
        \hline

	\end{tabular}
\end{table*}
% % \vspace{-4pt}

Following~\cite{ge2020selfpaced}, we conduct extensive experiments on multiple large-scale Re-ID benchmarks, including two real-world person datasets and one synthetic person dataset, as well as two real-world vehicle datasets and one synthetic vehicle dataset.
We evaluate our proposed method on both the mainstream real$\to$real adaptation tasks and the more challenging synthetic$\to$real adaptation tasks in person and vehicle Re-ID problems.
The details of these datasets are summarized in Table \ref{tab:dataset}.

% \vspace{-5pt}
\paragraph{Person Re-ID datasets}

Market-1501~\cite{market}, DukeMTMC-ReID~\cite{dukemtmc}, and MSMT17~\cite{wei2018person} are real-world person image datasets that are widely used in domain adaptive tasks.
MSMT17 includes more images that were captured in more challenging scenarios.
The synthetic PersonX database~\cite{sun2019dissecting} was constructed based on the Unity tool~\cite{riccitiello2015john} with manually designed challenges,
including random occlusion, resolution and illumination changes.

\paragraph{Vehicle Re-ID datasets}
To verify the generalization ability of our method on different kinds of objects,
we conduct experiments with the real-world
VeRi-776~\cite{liu2016deep}, VehicleID~\cite{liuhy2016deep}, and the synthetic VehicleX datasets.
% are typical datasets for evaluating vehicle re-ID.
VehicleX~\cite{naphade20204th} is also generated by the Unity engine~\cite{yao2019simulating,tang2019pamtri}
and further translated to the real-world style by SPGAN~\cite{deng2018image}.

\paragraph{Evaluation protocol}
In our experiments,
only ground-truth IDs of the source-domain datasets are provided for training. Experiments are conducted in line with the official evaluation protocol for each database. We adopt the widely used top-1/5/10 and mean Average Precision (mAP) as evaluation metrics. Moreover, following~\cite{ge2020mutual,zhai2020multiple,zheng2020exploiting}, the mean-Net is adopted for inference for both the baseline and our UCF method.

\subsection{Implementation Details}
\label{details}
We implement our framework in PyTorch \cite{pytorch}.
We adopt ResNet-50 \cite{he2016deep} as the backbone of the feature extractor and initialize the model with the parameters pre-trained on ImageNet \cite{deng2009imagenet}.
After Layer4 of the ResNet-50 model, we add one Generalized-Mean (GeM) pooling~\cite{radenovic2018fine} layer, one 1-D batch normalization \cite{57ioffe2015batch}
layer, and one L2-normalization layer. The L2-normalization layer produces 2048-dimensional feature vectors.
Following~\cite{luo2019bag}, we perform data augmentation via random erasing, cropping, and flipping.
For both source-domain pre-training and target-domain fine-tuning, we consistently construct a mini-batch with 64 person images of 16 identities. The person and vehicle images are resized to $256 \times 128$ and  $224 \times 224$ pixels, respectively. To achieve faster convergence, we adopt embeddings of cluster centroids to initialize the weights of the classifiers. The momentum coefficients in Eq. \ref{eq:update} and Eq. \ref{eq:ema} are set to 0.2 and 0.999, respectively. For DBSCAN, following~\cite{zhang2019self,ge2020selfpaced,li2020joint}, the hyper-parameter $d$ is set to 0.6 and the minimal number of neighbors in a core point is set to 4.  Following~\cite{ge2020selfpaced,zhong2019invariance}, the temperature $\tau$ in Eq. \ref{eq:loss_contrastive} is set as 0.05. The threshold $\alpha$ in hierarchical clustering is set to 0.0. The uncertainty threshold $\beta$ is set to 0.8. The ADAM method is adopted for optimization. The initial learning rate is set to 0.00035 and is decreased by multiplying by 0.1 on the 50-th epoch. The training lasts until the 80-th epoch.

\begin{table*}[htp]
\caption{Comparison with state-of-the-art UDA re-ID methods on real $\to$ real tasks
}
% \footnotesize
\scriptsize
\label{tab:sota-real2real}
\begin{center}
\begin{tabular}{l|c|p{1cm}<{\centering}p{1cm}<{\centering}p{1cm}<{\centering}p{1cm}<{\centering}|p{1cm}<{\centering}p{1cm}<{\centering}p{1cm}<{\centering}p{1cm}<{\centering}}
\hline
\multirow{2}{*}{Methods} & \multirow{2}{*}{Reference} & \multicolumn{4}{c|}{DukeMTMC-ReID$\to$Market-1501} & \multicolumn{4}{c}{Market-1501$\rightarrow$DukeMTMC-ReID} \\ \cline{3-10}
                         &                            & mAP      & top-1      & top-5       & top-10     & mAP      & top-1      & top-5      & top-10     \\ \hline
PUL \cite{fan2018unsupervised}                     & TOMM 2018                  & 20.5     & 45.5     & 60.7     & 66.7    & 16.4     & 30.0     & 43.4    & 48.5    \\
TJ-AIDL \cite{wang2018transferable}                 & CVPR 2018                  & 26.5     & 58.2     & 74.8     & 81.1    & 23.0     & 44.3     & 59.6    & 65.0    \\
SPGAN+LMP \cite{deng2018image}                & CVPR 2018                  & 26.7     & 57.7     & 75.8     & 82.4    & 26.2     & 46.4     & 62.3    & 68.0    \\
HHL \cite{zhong2018generalizing}                     & ECCV 2018                  & 31.4     & 62.2     & 78.8     & 84.0    & 27.2     & 46.9     & 61.0    & 66.7    \\
ECN \cite{zhong2019invariance}                     & CVPR 2019                  & 43.0     & 75.1     & 87.6     & 91.6    & 40.4     & 63.3     & 75.8    & 80.4    \\
PDA-Net \cite{li2019cross}                 & ICCV 2019                  & 47.6     & 75.2     & 86.3     & 90.2    & 45.1     & 63.2     & 77.0    & 82.5    \\
PCB-PAST \cite{zhang2019self}                & ICCV 2019                  & 54.6     & 78.4     & -        & -       & 54.3     & 72.4     & -       & -       \\
SSG \cite{fu2019self}                     & ICCV 2019                  & 58.3     & 80.0     & 90.0     & 92.4    & 53.4     & 73.0     & 80.6    & 83.2    \\
MMCL \cite{wang2020unsupervised}                 & CVPR 2020                 & 60.4     & 84.4     & 92.8     & 95.0    & 51.4     & 72.4     & 82.9    & 85.0    \\
ECN-GPP \cite{zhong2020learning}                 & TPAMI 2020                 & 63.8     & 84.1     & 92.8     & 95.4    & 54.4     & 74.0     & 83.7    & 87.4    \\
JVTC+ \cite{li2020joint} &ECCV 2020 &67.2 &86.8 &95.2 &97.1 &66.5 &80.4 &89.9 &92.2  \\
AD-Cluster \cite{zhai2020ad}              & CVPR 2020                  & 68.3     & 86.7     & 94.4     & 96.5    & 54.1     & 72.6     & 82.5    & 85.5    \\
MMT \cite{ge2020mutual}                     & ICLR 2020                  & 71.2     &87.7     &94.9     &96.9    &65.1     &78.0     &88.8    &92.5    \\
CAIL \cite{luo2020generalizing} &ECCV 2020 &71.5 &88.1 &94.4 &96.2 &65.2 &79.5 &88.3 &91.4 \\
NRMT \cite{zhao2020unsupervised} &ECCV 2020  &71.7 &87.8 &94.6 &96.5 &62.2 &77.8 &86.9 &89.5 \\
MEB-Net \cite{zhai2020multiple} &ECCV 2020 &76.0 &89.9 &96.0 &97.5 &66.1 &79.6 &88.3 &92.2 \\
SpCL \cite{ge2020self} &NeurIPS 2020 &76.7 &90.3 &96.2 &97.7 &68.8 &\underline{82.9} &90.1 &92.5 \\
Dual-Refinement \cite{dai2020dual} &TIP 2021 &78.0 &90.9 &96.4 &97.7 &67.7 &82.1 &90.1 &92.5 \\
UNRN \cite{zheng2020exploiting} &AAAI 2021 &78.1 &91.9 &96.1 &\underline{97.8} &69.1 &82.0 &\underline{90.7} &\underline{93.5}    \\
GLT \cite{zheng2021group} &CVPR 2021 &\underline{79.5} &\underline{92.2} &\underline{96.5} &\underline{97.8} &\underline{69.2} &82.0 &90.2 &92.8    \\ \hline
\multicolumn{2}{c|}{\textbf{Ours}} & \textbf{83.6} & \textbf{93.7} & \textbf{97.7} & \textbf{98.5} & \textbf{71.5} & \textbf{83.7} & \textbf{91.4} & \textbf{93.5} \\
                        \multicolumn{2}{c|}{Oracle} &82.7&94.1&97.9&98.8 &71.3 &84.5&92.2&94.2 \\
                        \hline \hline

\multirow{2}{*}{Methods} & \multirow{2}{*}{Reference} & \multicolumn{4}{c|}{Market-1501$\to$MSMT17} & \multicolumn{4}{c}{DukeMTMC-ReID$\to$MSMT17} \\ \cline{3-10}
 &  & mAP & top-1& top-5 & top-10 & mAP & top-1& top-5 & top-10 \\ \hline
ECN \cite{zhong2019invariance} & CVPR 2019 & 8.5 & 25.3 & 36.3 & 42.1 & 10.2 & 30.2 & 41.5 & 46.8 \\
SSG \cite{fu2019self} & ICCV 2019 & 13.2 & 31.6 & - & 49.6 & 13.3 & 32.2 & - & 51.2 \\
ECN-GPP \cite{zhong2020learning} & TPAMI 2020 & 15.2 & 40.4 & 53.1 & 58.7 & 16.0 & 42.5 & 55.9 & 61.5 \\
MMCL \cite{wang2020unsupervised} & CVPR 2020 & 15.1 & 40.8 & 51.8 & 56.7 & 16.2 & 43.6 & 54.3 & 58.9 \\
NRMT \cite{zhao2020unsupervised} &ECCV 2020 &19.8 &43.7 &56.5 &62.2 &20.6 &45.2 &57.8 &63.3 \\
CAIL \cite{luo2020generalizing} &ECCV 2020 &20.4 &43.7 &56.1 &61.9 &24.3 &51.7 &64.0 &68.9 \\
MMT \cite{ge2020mutual} & ICLR 2020 & 22.9 & 49.2 & 63.1 & 68.8 & 23.3 & 50.1 & 63.9 & 69.8 \\
JVTC+ \cite{li2020joint} &ECCV 2020 &25.1 &48.6 &65.3 &68.2 &27.5 &52.9 &\underline{70.5} &\underline{75.9} \\
SpCL \cite{ge2020self} &NeurIPS 2020 &\underline{26.8} &53.7 &65.0 &69.8 &26.5 &53.1 &65.8 &70.5 \\
Dual-Refinement \cite{dai2020dual} &TIP 2021 &25.1 &53.3 &66.1 &71.5 &26.9 &55.0 &68.4 &73.2 \\
UNRN \cite{zheng2020exploiting} & AAAI 2021 & 25.3 & 52.4 & 64.7 &69.7 & 26.2 & 54.9 & 67.3 & 70.6 \\
GLT \cite{zheng2021group} &CVPR 2021 &26.5  &\underline{56.6}  &\underline{67.5}  &\underline{72.0}  &\underline{27.7}  &\underline{59.5}  &70.1  &74.2  \\ \hline
\multicolumn{2}{c|}{\textbf{Ours}} & \textbf{34.8} & \textbf{66.1} & \textbf{76.6}  & \textbf{80.6} &\textbf{34.7} & \textbf{66.5} & \textbf{77.0} & \textbf{80.9} \\
                        \multicolumn{2}{c|}{Oracle} &45.1&74.5&84.8&88.0 &45.1&74.5&84.8&88.0 \\
\hline \hline

\multirow{2}{*}{Methods} & \multirow{2}{*}{Reference} & \multicolumn{4}{c|}{MSMT17 $\to$ Market-1501} & \multicolumn{4}{c}{MSMT17 $\to$ DukeMTMC-reID} \\ \cline{3-10}
 &  & mAP & top-1& top-5 & top-10 & mAP & top-1& top-5 & top-10 \\ \hline
CASCL \cite{wu2019unsupervised} &ICCV 2019 &35.5 &65.4 &80.6 &86.2 &37.8 &59.3 &73.2 &77.8 \\
MAR \cite{yu2019unsupervised} &CVPR 2019 &40.0 &67.7 &81.9 &87.3 &48.0 &67.1 &79.8 &84.2 \\
PAUL \cite{yang2019patch} &CVPR 2019 &40.1 &68.5 &82.4 &87.4 &53.2 &72.0 &82.7 &86.0 \\
DG-Net++ \cite{zou2020joint} &ECCV 2020 &64.6 &83.1 &91.5 &94.3 &58.2 &75.2 &73.6 &86.9 \\
D-MMD \cite{mekhazni2020unsupervised} &ECCV 2020 &50.8 &72.8 &88.1 &92.3 &51.6 &68.8 &82.6 &87.1 \\
MMT \cite{ge2020mutual} &ICLR 2020 &75.6 &89.3 &95.8 &97.5 &63.3 &77.4 &88.4 &91.7 \\
SpCL \cite{ge2020self} &NeurIPS 2020 &\underline{77.5} &\underline{89.7} &\underline{96.1} &\underline{97.6} &\underline{69.3} &\underline{82.9} &\underline{91.0} &\underline{93.0} \\ \hline
\multicolumn{2}{c|}{\textbf{Ours}} & \textbf{85.5} & \textbf{94.6} & \textbf{97.9} & \textbf{98.8}  & \textbf{71.5} & \textbf{84.1} & \textbf{91.6} & \textbf{93.6} \\
                        \multicolumn{2}{c|}{Oracle} &82.7&94.1&97.9&98.8 &71.3 &84.5&92.2&94.2 \\ \hline

\end{tabular}
\end{center}
\end{table*}

\begin{table*}[htp]
\caption{Comparison with state-of-the-art UDA re-ID methods on synthetic $\rightarrow$ real tasks}
% \footnotesize
\scriptsize
\label{tab:sota_synthetic2real}
\begin{center}
\begin{tabular}{l|c|cccc|cccc|cccc}
\hline
\multirow{2}{*}{Methods} & \multirow{2}{*}{Reference} & \multicolumn{4}{c|}{PersonX$\to$MSMT17} & \multicolumn{4}{c|}{PersonX$\to$Market1501} & \multicolumn{4}{c}{PersonX$\to$DukeMTMC-reID}\\ \cline{3-14}
                         &                            & mAP      & top-1      & top-5       & top-10     & mAP      & top-1      & top-5      & top-10   & mAP      & top-1      & top-5      & top-10   \\ \hline
MMT \cite{ge2020mutual} &ICLR 2020 & 17.7 & 39.1 & {52.6} & {58.5} & 71.0 & 86.5 & {94.8} & \underline{97.0} &60.1 &74.3 &86.5 &90.5\\
             SpCL \cite{ge2020selfpaced} &NeurIPS 2020 &\underline{22.7} & \underline{47.7} & \underline{60.0} & \underline{65.5} &\underline{73.8} & \underline{88.0} & \underline{95.3} & {96.9} &\underline{67.2} &\underline{81.8} &\underline{90.2} &\underline{92.6}\\\hline
             \multicolumn{2}{c|}{\textbf{Ours}} & \textbf{28.3} & \textbf{58.2} & \textbf{69.7} & \textbf{74.3} & \textbf{80.5} & \textbf{92.1} & \textbf{97.1} & \textbf{98.2} & \textbf{70.7} & \textbf{84.8} & \textbf{91.7} & \textbf{94.1}\\
                        \multicolumn{2}{c|}{Oracle}  &45.1&74.5&84.8&88.0 &82.7&94.1&97.9&98.8 &70.9 &84.5 &92.2 &94.2\\
 \hline
\end{tabular}
\end{center}
% \vspace{-3em}

\end{table*}

\begin{table*}[htp]
\caption{Performance comparisons with state-of-the-art UDA vehicle Re-ID methods
}
% \footnotesize
\scriptsize
\label{tab:sota_vehicle}
\begin{center}
\begin{tabular}{l|c|p{1cm}<{\centering}p{1cm}<{\centering}p{1cm}<{\centering}p{1cm}<{\centering}|p{1cm}<{\centering}p{1cm}<{\centering}p{1cm}<{\centering}p{1cm}<{\centering}}
\hline
\multirow{2}{*}{Methods} & \multirow{2}{*}{Reference} & \multicolumn{4}{c|}{VehicleID$\to$VeRi-776} & \multicolumn{4}{c}{VehicleX$\to$VeRi-776} \\ \cline{3-10}
                         &                            & mAP      & top-1      & top-5       & top-10     & mAP      & top-1      & top-5      & top-10     \\ \hline
MMT \cite{ge2020mutual} &ICLR 2020 & 35.3 & 74.6 & {82.6} & {87.0} & 35.6 & 76.0 & 83.1 & {87.4} \\
             SpCL \cite{ge2020selfpaced} &NeurIPS 2020 &\underline{38.9} & \underline{80.4} & \underline{86.8} & \underline{89.6}  & \underline{38.9} & \underline{81.3} & \underline{87.3} &\underline{90.0}\\\hline
             \multicolumn{2}{c|}{\textbf{Ours}} & \textbf{40.5} & \textbf{85.2} & \textbf{88.7} & \textbf{90.9} & \textbf{40.6} & \textbf{84.4} & \textbf{88.4} & \textbf{91.5} \\
            % \multicolumn{2}{c|}{Source Pretrain} &23.8&68.3&74.9&78.8 &26.2&65.2&76.1&82.0 \\
                        \multicolumn{2}{c|}{Oracle} &71.9 &93.6  &96.9  &98.3 &71.9 &93.6  &96.9  &98.3 \\
 \hline
\end{tabular}
\end{center}
% \vspace{-1.5em}

\end{table*}

\subsection{Comparison with State-of-the-Art Methods}

We compare the performance of UCF with state-of-the-art methods on multiple domain adaptation tasks, including real$\to$real and more challenging synthetic$\to$real tasks.
The performance of these methods is tabulated in Table \ref{tab:sota-real2real}, Table \ref{tab:sota_synthetic2real}, and Table \ref{tab:sota_vehicle}, respectively.
``Oracle'' stands for the Re-ID performance in the fully supervised setting.
It is clear that UCF significantly outperforms all state-of-the-art methods on both person and vehicle datasets with a plain ResNet-50 backbone.

\noindent\textbf{Results on real$\to$real UDA person Re-ID tasks}
We compare the performance of UCF with state-of-the-art methods on six UDA settings in Table~\ref{tab:sota-real2real}.
It is clear that UCF consistently outperforms existing approaches by large margins on all these benchmarks.
In particular, UCF outperforms MMT~\cite{ge2020mutual} by $12.4\%$, $6.4\%$, $11.9\%$, $11.4\%$, $9.9\%$, and $8.2\%$ in terms of mAP on these six tasks.
It is worth noting that both UCF and MMT adopt mean-Net during the training stage.
Moreover, UCF surpasses SpCL~\cite{ge2020selfpaced} by as much as 8.0\% and 12.4\% in terms of mAP and top-1 accuracy in the Market1501$\to$MSMT17 task, respectively.
% Although SpCL~\cite{ge2020selfpaced} effectively still uses source domain data during the training process, and our method is source-free.
Finally, UCF significantly outperforms one very recent method named GLT \cite{zheng2021group} by $8.3\%$ and $7.0\%$ in terms of the mAP accuracy, for Market-1501$\to$MSMT17 and DukeMTMC-ReID$\to$MSMT17 tasks, respectively. The above experimental results clearly demonstrates the effectiveness of UCF.

UCF also significantly bridges the gap between the unsupervised and fully-supervised settings. For example, UCF achieves 94.6\% top-1 accuracy and 85.5\% mAP on the MSMT17$\to$Market1501 task, meaning that it surpasses the the performance of ``Oracle'' on the Market-1501 database by 0.5\% in top-1 accuracy and 2.8\% in mAP, respectively.
In addition to the reliable pseudo labels generated by UCF, another possible reason is that MSMT17 is larger than Market1501;
therefore, supervised pre-training on MSMT17 provides better model initialization before domain adaptation.

\noindent\textbf{Results on synthetic$\to$real UDA person Re-ID tasks}
\label{person reid}
Compared with the real$\to$real UDA re-ID tasks, the synthetic$\to$real UDA tasks are usually more challenging due to the dramatic domain gap.
As shown in Table \ref{tab:sota_synthetic2real}, UCF outperforms state-of-the-art methods by large margins.
For example, UCF beats the SpCL~\cite{ge2020self} method by 4.1\% in terms of top-1 accuracy and 6.7\% in terms of mAP on the PersonX$\to$Market-1501 task.
It is also worth noting that the performance of UCF in synthetic$\to$real tasks still exceeds that of SpCL in real$\to$real tasks.
Specifically, UCF achieves 92.1\% top-1 accuracy and 80.5\% mAP on the PersonX$\to$Market1501 task, which outperform SpCL~\cite{ge2020self} on the MSMT17$\to$Market-1501 task by 2.4\% in terms of top-1 accuracy and 3.0\% in terms of mAP.

Although these results are promising, there is still a clear gap between UCF and ``Oracle'' on large-scale datasets such as MSMT17.
This motivates us to develop more robust clustering and pseudo label generation methods in the future.

\noindent\textbf{Results on vehicle Re-ID datasets}
As Table \ref{tab:sota_vehicle} shows, the performance of UCF surpasses that of SpCL by 4.8\% in top-1 accuracy and 1.6\% in mAP on the VehicleID$\to$VeRi-776 task.
Moreover, UCF outperforms SpCL by 3.1\% in top-1 accuracy and 1.7\% in mAP on the VehicleX$\to$VeRi-776 task.
These experimental results further demonstrate the effectiveness of UCF for object Re-ID.

\subsection{Ablation Studies}
\label{sec:ablation}

 \begin{table*}[t]
    \scriptsize
    % \footnotesize
    % \small
    % \tiny
    \centering
    \caption{Ablation studies on each key component of UCF}
    \begin{center}
    \begin{tabular}{P{2.0cm}|C{0.8cm}C{0.8cm}C{0.8cm}C{0.9cm}|C{0.8cm}C{0.8cm}C{0.8cm}C{0.9cm}}
                \hline
            \multicolumn{1}{c|}{\multirow{2}{*}{Methods}} & \multicolumn{4}{c|}{Market1501$\to$MSMT17} &

            \multicolumn{4}{c}{PersonX$\to$MSMT17} \\
            \cline{2-9}
            \multicolumn{1}{c|}{} & mAP & top-1 & top-5 & top-10 & mAP & top-1 & top-5 & top-10 \\
            \hline
                     \multicolumn{1}{l|}{Source Pretrain} &8.4&22.6&32.9&38.1 &2.7&8.8&14.8&18.3 \\
                                  \hline

             \multicolumn{1}{l|}{Cross-entropy loss} & 28.5 & 57.5 & {69.6} & {74.3} &23.5 & 51.4 & 64.3 & 69.5\\
            %  \multicolumn{1}{l|}{Contrastive loss} &{25.7} & {53.9} & {65.2} & {69.9}  &22.4  &48.8 &60.3 &64.9\\
             \multicolumn{1}{l|}{Contrastive loss} &{26.8} & {55.0} & {66.5} & {71.5}  &22.4  &48.8 &60.3 &64.9\\
             \multicolumn{1}{l|}{Strong baseline} &{31.6} & {61.7} &72.3  & 76.4  & 26.2  & 55.0  & 67.0  & 71.6\\
            \hline
                         \multicolumn{1}{l|}{Baseline \textit{w/} HC} &33.8 &63.4  &74.5  &78.9  &27.6  &57.1 & 68.8 &73.3\\
                         \multicolumn{1}{l|}{Baseline \textit{w/} UCIS} &33.3 &64.6  &75.0  &79.0   & 27.6 & 57.4 & 68.6  &73.3\\
\multicolumn{1}{c|}{\textbf{Ours(full)}} & \textbf{34.8} & \textbf{66.1} & \textbf{76.6} & \textbf{80.6} & \textbf{28.3} & \textbf{58.2} & \textbf{69.7} & \textbf{74.3} \\
            \hline
            % \hline
            \end{tabular}
        \end{center}
        \label{tab:ablation}
        % \vspace{-20pt}
    \end{table*}

 \begin{table*}[t]
    \scriptsize
    % \footnotesize
    % \small
    % \tiny
    \centering
    \caption{Performance comparison between our class-level contrastive loss and instance-level contrastive loss}
    \begin{center}
    \begin{tabular}{P{2.0cm}|C{0.8cm}C{0.8cm}C{0.8cm}C{0.9cm}|C{0.8cm}C{0.8cm}C{0.8cm}C{0.9cm}}
                \hline
            \multicolumn{1}{c|}{\multirow{2}{*}{Methods}} & \multicolumn{4}{c|}{Market1501$\to$MSMT17} &

            \multicolumn{4}{c}{PersonX$\to$MSMT17} \\
            \cline{2-9}
            \multicolumn{1}{c|}{} & mAP & top-1 & top-5 & top-10 & mAP & top-1 & top-5 & top-10 \\
            \hline
             \multicolumn{1}{l|}{Instance-level contrastive loss} &{24.3} & {52.5} & {64.1} & {69.1}  &19.1  &45.3 &56.2 &61.4\\
             \multicolumn{1}{l|}{Class-level contrastive loss (ours)} &{26.8} & {55.0} & {66.5} & {71.5}  &22.4  &48.8 &60.3 &64.9\\

            \hline
            % \hline
            \end{tabular}
        \end{center}
        \label{tab:c_loss}
        % \vspace{-20pt}
    \end{table*}

We systematically investigate the effectiveness of each key component of UCF: namely, the strong baseline, hierarchical clustering (HC), and uncertainty-aware collaborative instance selection (UCIS), respectively. Experiments are conducted on real$\to$real and more challenging synthetic$\to$real adaptation tasks, specifically Market1501$\to$MSMT17 and PersonX$\to$MSMT17. The results are summarized in Table \ref{tab:ablation}. ``Source Pretrain'' represents the Re-ID model trained in the source domain and tested directly in the target domain.

\noindent\textbf{Effectiveness of the strong baseline}
We build our baseline with the cross-entropy loss and our new contrastive loss, both of which are described in Section \ref{sec:baseline}. We first evaluate the performance when only classification loss or contrastive loss is used.
As shown in Table \ref{tab:ablation}, the two settings achieve 28.5\% and 26.8\% mAP respectively for the Market1501$\to$MSMT17 task.
In addition, as shown in Table \ref{tab:c_loss}, our new contrastive loss outperforms the conventional instance-level contrastive loss by 2.5\% and 3.3\% mAP on the two UDA tasks, respectively.
When the two loss functions are used together, we obtain a strong baseline.
For example, compared with ``Source Pretrain'' in Table \ref{tab:ablation}, our baseline promotes the top-1 accuracy
by 39.1\% and 46.2\%, as well as mAP by 23.2\% and 23.5\%, on the two UDA tasks, respectively.
 These results prove that our baseline is simple but effective.

%\vspace{-5pt}
\noindent\textbf{Effectiveness of the hierarchical clustering}
 Compared with our baseline, the hierarchical clustering method consistently yields performance gains. For example, ``Baseline \textit{w/} HC'' outperforms the baseline in terms of top-1 accuracy by 1.7\% and 2.1\%, as well as mAP by 2.2\% and 1.4\%, on Market1501$\to$MSMT17 and PersonX$\to$MSMT17 tasks, respectively. This is because the hierarchical clustering  improves the quality of pseudo-labels, meaning that the deep model can learn more discriminative features.

 \noindent\textbf{Effectiveness of the uncertainty-aware collaborative instance selection}
When the baseline is equipped with the UCIS module, the performance of both UDA tasks is promoted. In particular, UCIS improves the top-1 accuracy of the baseline by 2.9\% and 2.4\%, as well as mAP by 1.7\% and 1.4\%, on the two tasks, respectively. The above results demonstrate the necessity of reducing the impact of noisy labels, as well as the effectiveness of our method.

\noindent\textbf{Effectiveness of the UCF framework}
Finally, with both the HC and UCIS modules, our full model achieves better performance than using either of the modules alone. The above comparisons justify the effectiveness of each key component in our framework.

Furthermore, we test the performance of SpCL~\cite{ge2020self} based on our strong baseline.
We equip SpCL with a hybrid memory to save target-domain cluster centroids and target-domain un-clustered instance features.
Experimental results are summarized in Table \ref{tab:samebaseline}.
It is shown that UCF still outperforms SpCL by 2.6\% and 1.5\% in terms of top-1 accuracy and mAP on Market1501$\to$MSMT17 task, respectively.
The above experimental results justify the effectiveness of UCF.

\begin{figure}
    \begin{minipage}[t]{0.48\textwidth}
        \centering
        \includegraphics[width=1.0\textwidth]{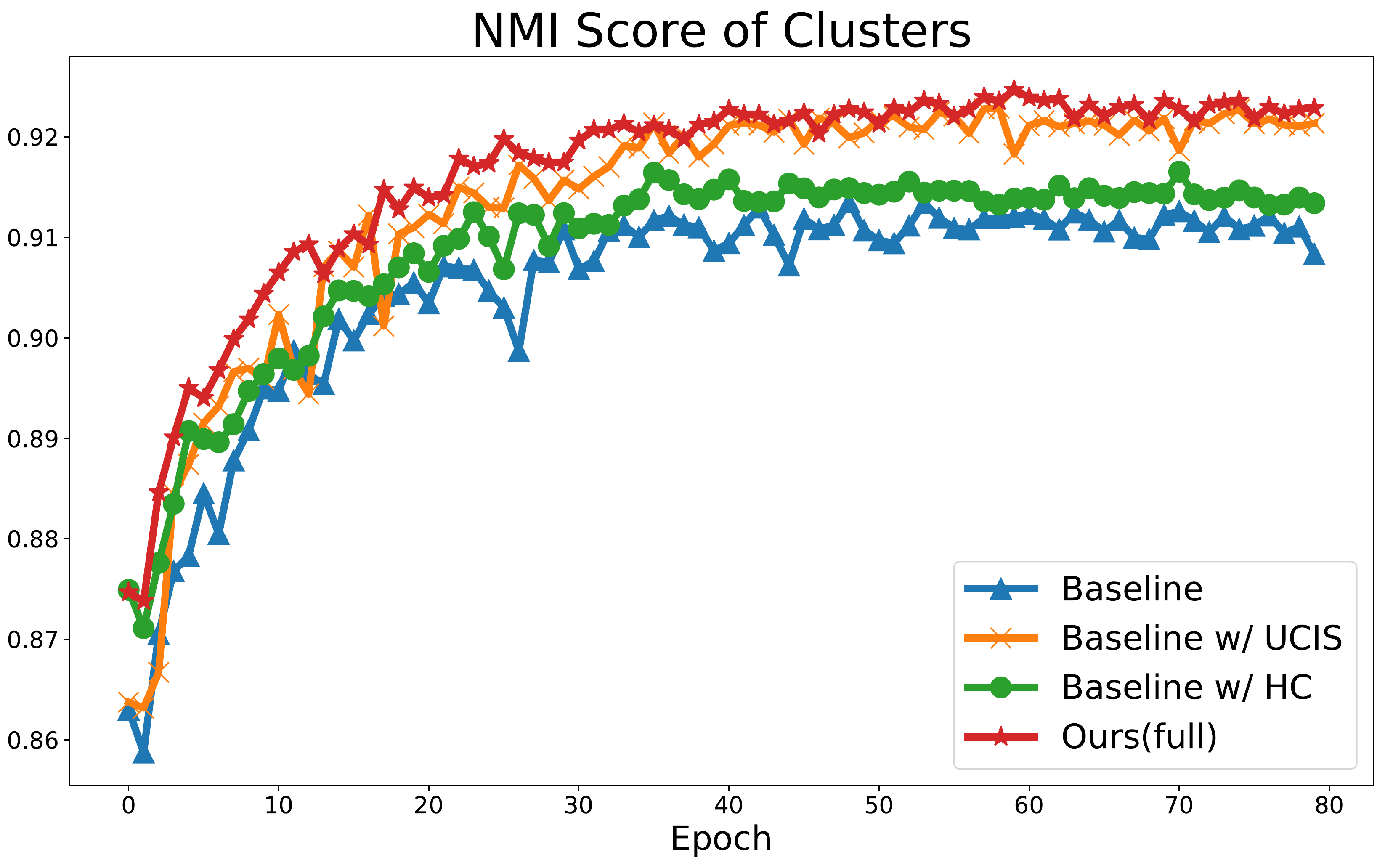}
        \caption{Comparisons on the Normalized Mutual Information (NMI) scores of clusters during the training process on the Market1501→MSMT17 task.}
        \label{fig:nmi}
    \end{minipage}\;\;\;\;
    \begin{minipage}[t]{0.48\textwidth}
        \centering
        \includegraphics[width=1.0\textwidth]{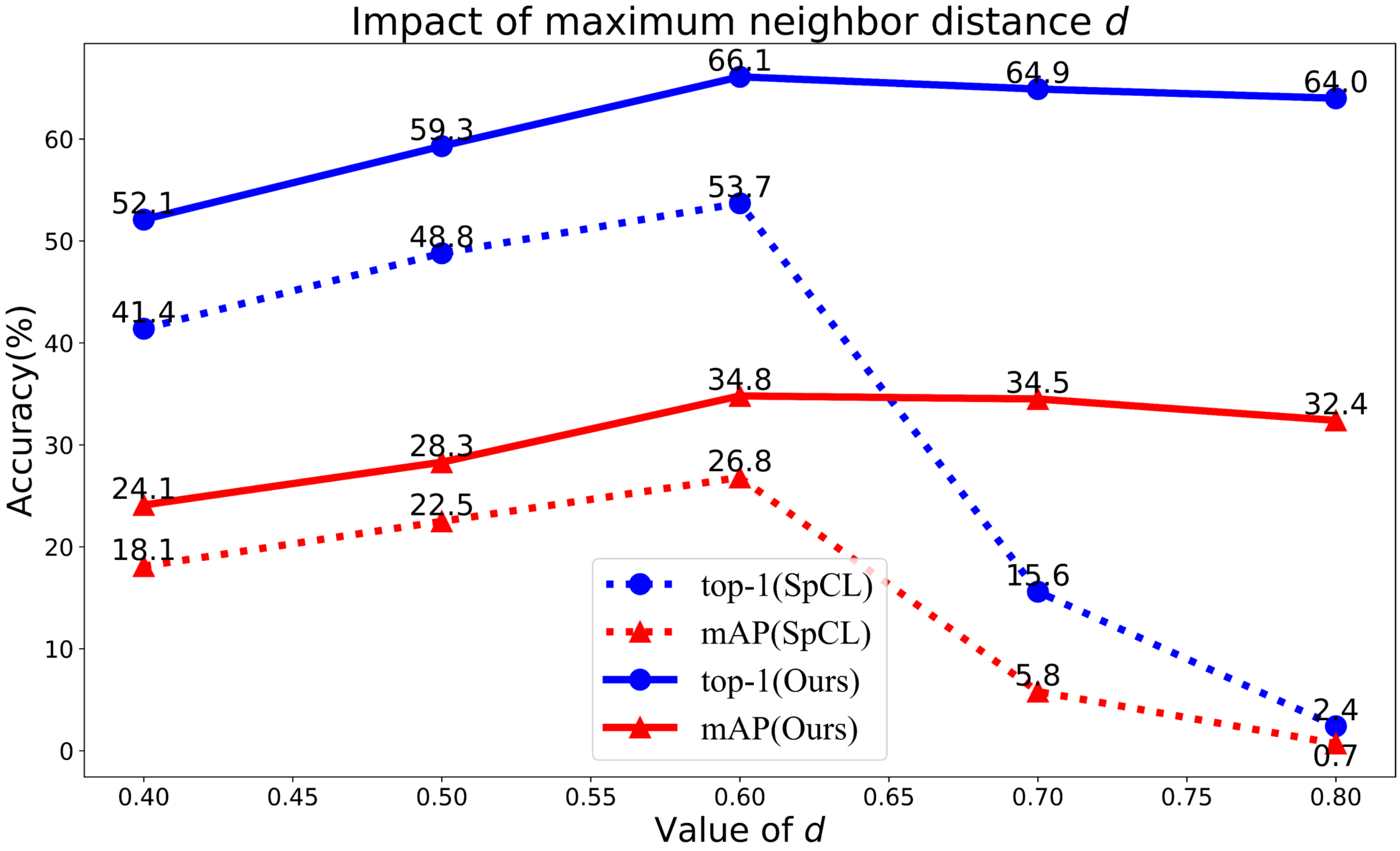}
        \caption{Performance comparison between UCF and SpCL~\cite{ge2020selfpaced} with different values of hyper-parameter $d$.}
        \label{fig:d}
    \end{minipage}
\end{figure}

\noindent\textbf{Analysis of the quality of pseudo labels}
 In Fig. \ref{fig:nmi}, we illustrate the improvement in the quality of pseudo labels. Following SpCL~\cite{ge2020selfpaced}, we illustrate the Normalized Mutual Information (NMI) \cite{mcdaid2011normalized} scores of clusters during training on the Market1501$\to$MSMT17 task. NMI \cite{mcdaid2011normalized} is an index that measures the accuracy of the clustering results. It can accordingly be observed that, compared with the baseline, the quality of the pseudo-labels is significantly improved when the proposed techniques are applied.

 \begin{table*}[t]
    \scriptsize
    % \footnotesize
    % \small
    % \tiny
    \centering
    \caption{Performance comparison between UCF and SpCL~\cite{ge2020selfpaced} with our strong baseline}
    \begin{center}
    \begin{tabular}{P{2.0cm}|C{0.8cm}C{0.8cm}C{0.8cm}C{0.9cm}|C{0.8cm}C{0.8cm}C{0.8cm}C{0.9cm}}
                \hline
            \multicolumn{1}{c|}{\multirow{2}{*}{Methods}} & \multicolumn{4}{c|}{Market1501$\to$MSMT17} &

            \multicolumn{4}{c}{PersonX$\to$MSMT17} \\
            \cline{2-9}
            \multicolumn{1}{c|}{} & mAP & top-1 & top-5 & top-10 & mAP & top-1 & top-5 & top-10 \\
            \hline
            \multicolumn{1}{c|}{Strong baseline} &{31.6} & {61.7} &72.3  & 76.4  & 26.2  & 55.0  & 67.0  & 71.6\\
             \multicolumn{1}{l|}{Strong baseline+SpCL~\cite{ge2020self}} &{33.3} & {63.5} & {74.0} & {78.6}  &27.3  &56.8 &68.0 &73.4\\
             \multicolumn{1}{c|}{\textbf{Ours(full)}} & \textbf{34.8} & \textbf{66.1} & \textbf{76.6} & \textbf{80.6} & \textbf{28.3} & \textbf{58.2} & \textbf{69.7} & \textbf{74.3} \\

            \hline
            % \hline
            \end{tabular}
        \end{center}
        \label{tab:samebaseline}
        % \vspace{-20pt}
    \end{table*}

\subsection{Parameter Analysis}

We tune the hyper-parameters on the  Market-1501$\to$MSMT17 task, then directly apply the chosen hyper-parameters to all the other tasks.

\noindent\textbf{Maximum neighbor distance $d$ for DBSCAN.}
DBSCAN is one of the most popular clustering algorithms in the UDA Re-ID literature. For DBSCAN, the maximum neighborhood distance $d$ is an important hyperparameter. As demonstrated in Fig. \ref{fig:d}, we find that the value of $d$ may considerably affect the performance of state-of-the-art methods. In particular, a larger value of $d$ may result in a dramatic performance drop; this is because the pseudo-labels will contain more noise as the value of $d$ increases. In comparison, the performance of UCF is significantly more robust.
% For example, the performance drop is only about 2.1\% when the value of d increases from 0.6 to 0.8; t
This is because UCF successfully improves the clustering quality and removes samples with unreliable pseudo-labels.

\begin{figure}
    \begin{minipage}[h]{0.48\textwidth}
        \centering
        \includegraphics[width=1.0\textwidth]{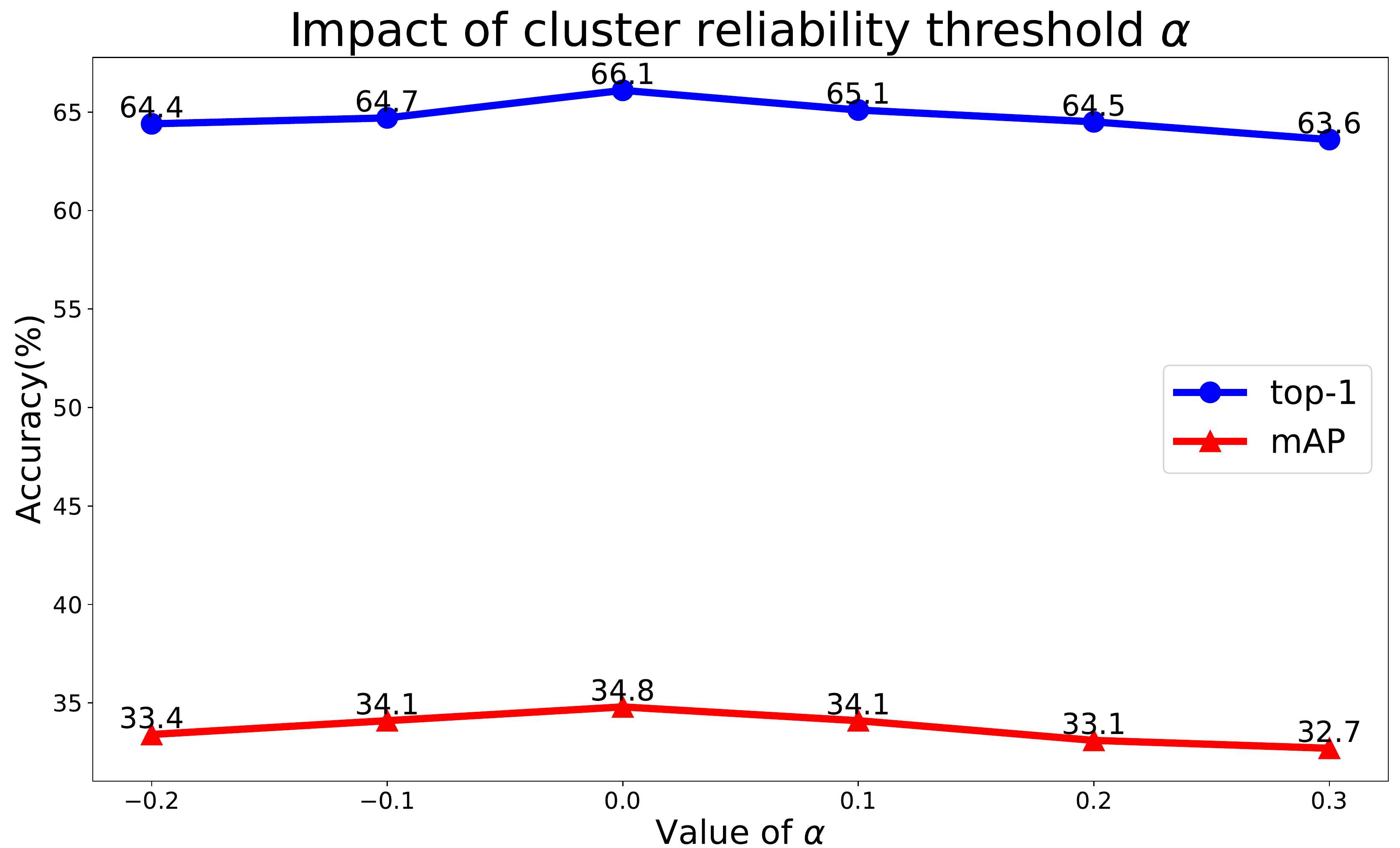}
        \caption{Performance of UCF with different values of $\alpha$.}
        \label{fig:alpha}
    \end{minipage}\;\;\;\;
    \begin{minipage}[h]{0.48\textwidth}
        \centering
        \includegraphics[width=1.0\textwidth]{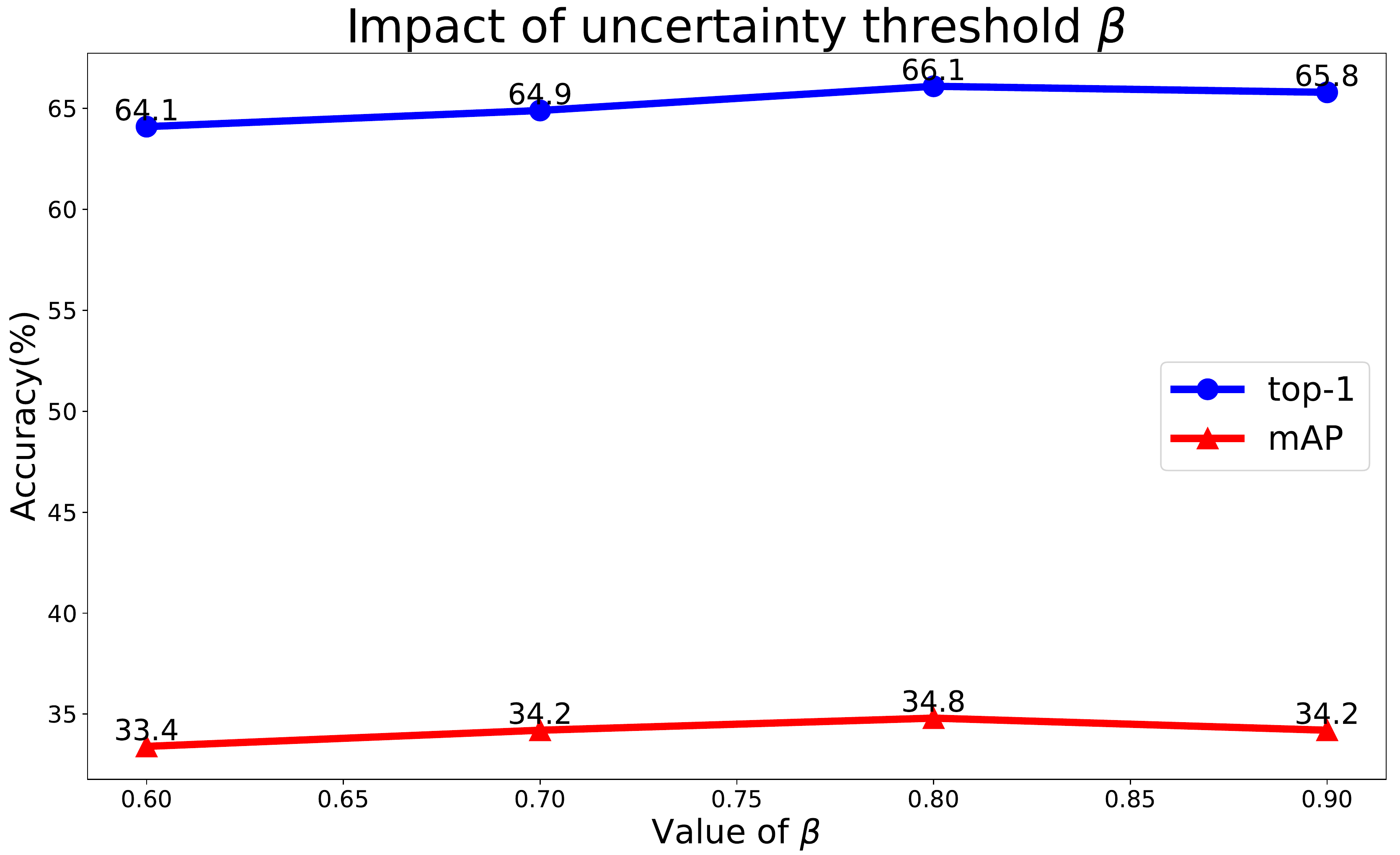}
        \caption{Performance of UCF with different values of $\beta$.}
        \label{fig:beta}
    \end{minipage}
    \end{figure}

\noindent\textbf{Cluster reliability threshold $\alpha$ for hierarchical clustering}
$\alpha$ is a threshold on $\mathcal{S}(\mathcal{I}_k)$. According to the definition of $\mathcal{S}(\mathcal{I}_k)$, a negative value of $\mathcal{S}(\mathcal{I}_k)$ means that intra-class distance surpasses inter-class distance. This usually indicates unreliable clustering from an object Re-ID perspective. As demonstrated in Fig. \ref{fig:alpha}, our framework achieves the optimal performance when $\alpha$ is set to 0.0 on the MSMT17$\to$Market-1501 task, which is consistent with our above analysis. When $\alpha$ is larger than 0.0, the top-1 accuracy and mAP gradually decrease. This is because some reliable clusters will be forced to be decomposed, resulting in more noisy pseudo-labels and therefore performance degradation.

\noindent\textbf{Uncertainty threshold $\beta$ for collaborative instance selection}
As described in Section \ref{Methodology}, we require an uncertainty threshold $\beta$ to select samples with reliable pseudo-labels. In Fig. \ref{fig:beta}, we investigate the effect of different values of $\beta$. As can be seen from Fig. \ref{fig:beta}, the performance of UCF is generally robust to the value of $\beta$ while the best performance is achieved when $\beta$ is set to 0.8. The performance of UCF reduces when $\beta$ is set to a smaller value, such as 0.6; this may be because samples with noisy pseudo-labels cannot be identified when the threshold is low.

\begin{figure}[t]
\centering
\includegraphics[width=1.0\linewidth]{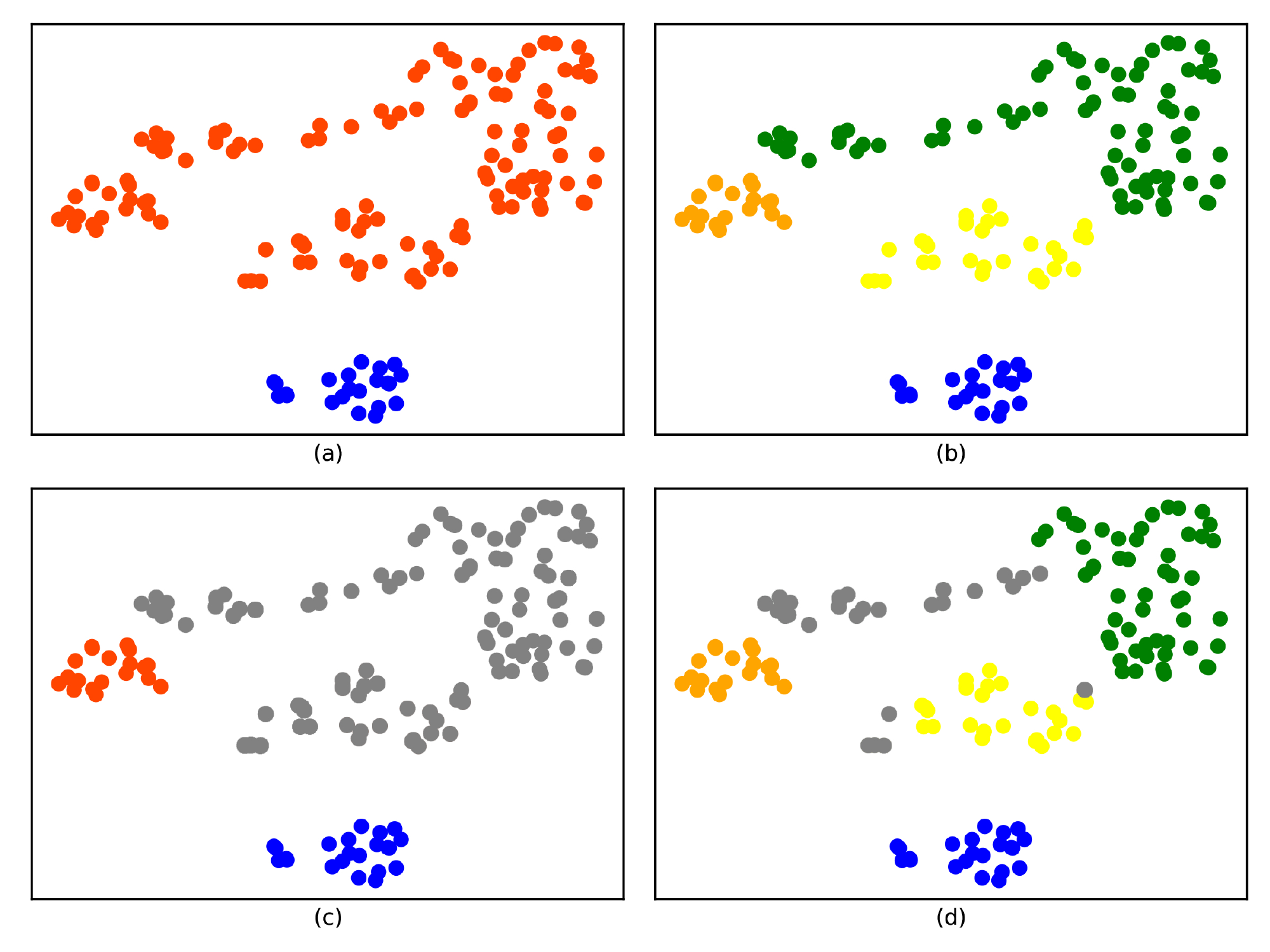}
\caption{Visualization using t-SNE \cite{van2008visualizing} for the clustering results on the target domain by four models: (a) ``Baseline'', (b) ``Baseline $w/$ HC'', (c) ``Baseline $w/$ UCIS'', and (d) the ``UCF model''. Different clusters are represented using different colors. Gray denotes outliers. (Best viewed in color.)}
%\vspace{-10pt}
\label{fig:vis}
\end{figure}

\subsection{Qualitative Comparisons}
In Fig. \ref{fig:vis}, we utilize t-SNE \cite{van2008visualizing} to visualize the clustering results by ``Baseline'', ``Baseline $w/$ HC'', ``Baseline $w/$ UCIS'', and the ``UCF'' model for the Market-1501$\to$MSMT17 task, respectively.  We have the following observations.

First, as illustrated in Fig. \ref{fig:vis}(a), due to the limited discriminative power of the Re-ID model in the target domain, many visually similar images may be grouped into the same cluster. The size of such clusters is often large.
Second, as illustrated in Fig. \ref{fig:vis}(b), when the proposed hierarchical clustering (HC) method is utilized, the unreliable clusters in Fig. \ref{fig:vis}(a) are decomposed into multiple smaller ones.
Third, as shown in Fig. \ref{fig:vis}(c), the uncertainty-aware collaborative instance selection (UCIS) method identifies instances with unreliable pseudo labels, which are represented using the gray color in the figure.
Finally, combining UCIS and HC can achieve the best clustering results, which proves that the two modules are complementary. The above visualization results are consistent with the results in the experimentation section.

\section{Conclusion}

In this work, we propose an uncertainty-aware clustering framework (UCF) to tackle the problem of noisy pseudo labels in clustering-based UDA object Re-ID tasks. UCF handles the label noise problem on two levels. First, a novel hierarchical clustering scheme is proposed to promote the clustering quality; second, an uncertainty-aware collaborative instance selection method is introduced to select images with reliable labels for model training. These two techniques significantly relieve the noise in pseudo-labels and consequently improve the quality of deep feature learning.
Our UCF method significantly outperforms state-of-the-art object Re-ID methods on many domain adaptation tasks.

\bibliographystyle{splncs04}
\small
\bibliography{neurips_2021}
\normalsize

\end{document}